\documentclass[letterpaper, 10 pt, conference]{ieeeconf} 
\IEEEoverridecommandlockouts                              % This command is only needed if 
\usepackage{graphicx}
\usepackage{times} % assumes new font selection scheme installed
\usepackage{color} % for \textcolor (as in \todo command)
\usepackage[bookmarks=true]{hyperref} % internal and external hyperlinks
\usepackage{multicol} % two-column format
\usepackage{verbatim}
\usepackage{verbatimbox}

{\small\verbatim}%
{\endverbatim}

\usepackage{outlines}
\usepackage{amsmath, amssymb, bm}
\usepackage{epsfig}
\usepackage{graphics}
\usepackage{subcaption}
\usepackage{sidecap}
\usepackage{multirow}
\usepackage{amssymb}
\usepackage{color}
\usepackage{url}
\usepackage{hyperref}
\usepackage{array}
\usepackage{balance}
\usepackage[linesnumbered,vlined,ruled]{algorithm2e}
\usepackage{flushend}
\usepackage{xspace}

\newcommand{\fig}[1]{Fig.~\ref{#1}}

\newcommand{\sect}[1]{Sec.~\ref{#1}}

\newcommand{\eg}[0]{{\em e.g.,~}}
\newcommand{\ie}[0]{{\em i.e.,~}}

\title{\LARGE \bf
End-User Programming of Low- and High-Level Actions\\for Robotic Task Planning
}

\author{Ying Siu Liang, Damien Pellier,
Humbert Fiorino, and Sylvie Pesty%$^{1}$
\thanks{%$^{1}$
Y.S. Liang, D. Pellier, H. Fiorino, and S. Pesty are with the
    Univ. Grenoble Alpes, LIG, F-38000 Grenoble, France
    {\tt\small \{liangyi, pellierd, fiorinoh, pestys\}@univ-grenoble-alpes.fr}}%
}

\begin{document}

\maketitle
\thispagestyle{empty}
\pagestyle{empty}

%%%%%%%%%%%%%%%%%%%%%%%%%%%%%%%%%%%%%%%%%%%%%%%%%%%%%%%%%%%%%%%%%%%%%%%%%%%%%%%%
\begin{abstract}
Programming robots for general purpose applications is extremely challenging due to the great diversity of end-user tasks ranging from manufacturing environments to personal homes.
Recent work has focused on enabling end-users to program robots using Programming by Demonstration. 
However, teaching robots new actions from scratch that can be reused for unseen tasks remains a difficult challenge and is generally left up to robotic experts.
We propose iRoPro, an interactive Robot Programming framework that allows end-users to teach robots new actions from scratch and reuse them with a task planner.
In this work we provide a system implementation on a two-armed Baxter robot that  
(i) allows simultaneous teaching of low- and high-level actions by demonstration, 
(ii) includes a user interface for action creation with condition inference and modification, and
(iii) allows creating and solving previously unseen problems using a task planner for the robot to execute in real-time.
We evaluate the generalisation power of the system on six benchmark tasks and show how taught actions can be easily reused for complex tasks.
We further demonstrate its usability with a user study (N=21), where users completed eight tasks to teach the robot new actions that are reused with a task planner.
The study demonstrates that users with any programming level and educational background can easily learn and use the system.
\end{abstract}

%%%%%%%%%%%%%%%%%%%%%%%%%%%%%%%%%%%%%%%%%%%%%%%%%%%%%%%%%%%%%%%%%%%%%%%%%%%%%%%%
\section{INTRODUCTION}
Despite the ongoing advances in Robotics and A.I., it is extremely challenging to pre-program robots for specific end-user applications.
Instead of developing robots for domain-specific tasks, a more flexible solution is to have robots learn new actions directly from end-users and let them customise the robot for their specific application.
Programming by Demonstration (PbD) \cite{billard2008robot} has been used to allow end-users to teach robots actions in an intuitive way by taking demonstrations as input and inferring a policy for the task.
However, PbD solutions usually require users to teach robots an action sequence %, which they believe is the solution 
to achieve a certain goal.
If the goal changes, the user has to teach the robot a new sequence.

Consider the Tower of Hanoi% \cite{lucas1884}
, a puzzle consisting of three pegs and a number of differently-sized disks, stacked on one peg in descending order, with the largest peg at the bottom.
The goal is to move the entire stack from one peg to another, by moving one disk at a time, and only to a larger disk or an empty peg. 
The solution is different depending on the given number of disks.
If we want to teach a robot to solve this problem% for any given number of disks
, it would be infeasible to demonstrate the solution each time.
%as it is different for each case.
% The action sequence to solve this problem always consists of a combination of the same primitive action, namely, to move a disk from one peg to another.
A more efficient approach would be to teach the robot the primitive action of moving a disk, associate rules or \textit{conditions} to this action (\eg smaller disks can only be placed on top of larger ones), and have the robot generate an optimal solution using a task planner \cite{ghallab2004automated}.

  \begin{figure}[t]
   \centering
\includegraphics[width=0.98\linewidth]{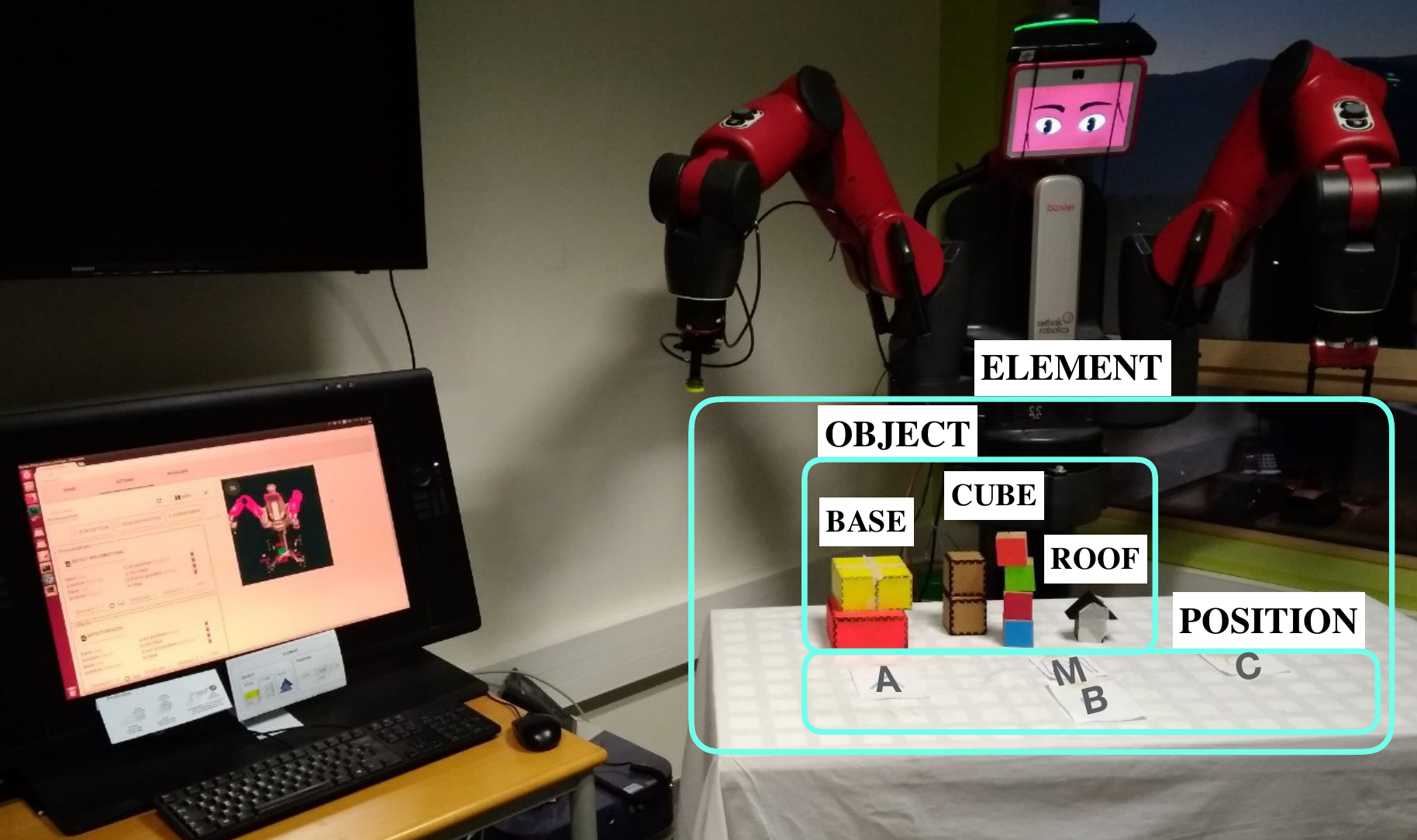}
   \caption{Users programmed the Baxter robot via a graphical interface to manipulate objects (shown with their type hierarchies) in the task domain.}
   \label{fig:dispositif}
  \end{figure}
Our research argues for teaching robots primitive actions, instead of entire action sequences, and delegating the logical reasoning process of finding a solution to task planners.
Even though task planners are generally used by domain-experts, we have previously shown that users with little to no programming experience can easily learn and use symbolic planning languages \cite{liang2017evaluation}.
%While previous work has addressed learning actions for symbolic task planning (\citet{abdo2013learning}), %and evaluated using the Wizard-of-Oz technique,
% While a robot programming framework has been proposed in theory,
% enabling end-users to program robots in this manner is non-trivial. 
Based on the obtained results, this work presents iRoPro, an interactive Robot Programming system (\sect{sec:approach}).
It is a working end-to-end system that allows efficient programming of both \textit{how} an action is performed (low-level) and \textit{when} it can be applied (high-level), while generalising both aspects to previously unseen scenarios. 
% Furthermore, it should provide an intuitive interface to reuse the programmed actions together with a task planner that is generally handled by domain experts.
% In this work we address this challenge of creating this system.
%Therefore we argue for enabling end-users to teach robots reusable actions and that users with or without programming experience are able to easily program robots to complete previously unseen tasks.
% The main contribution of this paper is iRoPro, an end-to-end system for teaching robots primitive actions to be reused with a task planner (\fig{fig:overview}).
% As described in \sect{sec:approach}, the robot simultaneously learns the low- and high-level action representations from a single user demonstration.
% A generalised action is inferred that can be reused directly with a task planner to solve problems that go beyond the initial demonstration.
% Thus, we enable end-users to program robots from scratch, without writing code, therefore maximising the generalisability of taught actions with minimum programming effort.
We implement the system on a Baxter robot and developed a graphical interface that allows users to teach new actions by kinesthetic demonstration, modify their conditions and define previously unseen problems to solve with a task planner (\sect{sec:system}).
We demonstrate our system's capability to generalise primitive actions on six benchmark tasks that are programmed and executed on the robot (\sect{sec:syseval}).
We empirically investigate the system's usability and validate its intuitiveness through a study with users of different educational backgrounds and programming levels
%and perform an analysis on both quantitative and qualitative data 
(\sect{sec:quanteval}).
To better understand user teaching strategies, we split participants into two control groups, with and without automatic condition inference, and showed that users in both groups can easily learn and use the system.
Finally, we discuss limitations and possible extensions to further increase the system's generalisability (\sect{sec:discussions}).
% In summary, the contributions of this paper are:
% \begin{enumerate}
% \item We provide an end-to-end system to teach the robot new actions from single demonstrations
% \item We analyse how end-users with different programming experience use this system.
% \item %The system translates user input into PDDL which is sent to a task planner to solve user-specified goals. 
% We analyse how users without experience in task planning create and solve planning problems.
% \item We augment our system with an inference model for generating parameters, preconditions, and effects (\sect{sec:inference}) and analyse user strategies with and without condition inference.
% \end{enumerate}

\section{RELATED WORK}
\label{sec:relatedwork}
Our work relates to several topics explored in previous research.
End-user robot programming has been addressed previously for industrial robots to be programmed by non-robotics domain experts, where users specify and modify existing plans for robots to adapt to new scenarios \cite{stenmark2017simplified}.
%CoSTAR \citet{paxton2017costar} use Behaviour Trees represent task plans that can be easily adapted by non-robotic end-users for a variety of tasks.
In our work we argue for the use of task planners to automatically generate plans for new scenarios.

Previous work has addressed knowledge engineering tools for constructing planning domains but usually require PDDL experts %(PDDL Studio \cite{plch2012inspect}), 
or common knowledge in software engineering (\eg itSIMPLE \cite{vaquero2013itsimple}).
%Integrating task planning with robotic systems has been addressed with
%the ROSPlan framework  which provides a collection of tools to integrate task planning in ROS systems.
There has been previous work on integrating task planning with robotic systems \cite{cashmore2015rosplan} and learning preconditions and effects of actions to be used in planning \cite{konidaris2018fromSkills}.
However, the robot is usually provided with a fixed set of low-level motor skills.
%It assumes that low-level actions have already been programmed and can be called directly after mapping them to PDDL actions.
We do not provide the robot with any predefined actions but allow users to teach both low- and high-level actions from scratch.

Programming by demonstration \cite{billard2008robot} has been commonly applied to allow end-users to teach robots new actions by demonstration.
Recent work has focused on mobile and industrial manipulators \cite{stenmark2017simplified}
%\cite{huang2017code3,stenmark2017simplified}
and learning from single demonstrations \cite{alexandrova2014robot}. %\cite{perez2017c,wu2010towards,yu2018one}.
%\citet{stenmark2017simplified} uses assembly tasks to pick and stack lego blocks and compares the performance of non-experts with reusable tasks (N=21).
Alexandrova et al. created an end-user programming framework with an interactive action visualisation allowing the user to teach new actions from single demonstrations but do not reuse them with a task planner.
%The system is evaluated on 12 benchmark tasks as well as a user study (N=10) for box closing tasks, stacking cups, and putting objects into a box.

%In \citet{konidaris2018fromSkills} the robot learns symbolic representations for high-level task planning by gathering data from 167 motor skills executions.
%The authors mention the drawback of unnecessary learning and the need to learn a representation that generalises across related motor skills.
Most closely related to our approach is the work by Abdo et al. \cite{abdo2013learning} where manipulation actions are learned from kinesthetic demonstrations and reused with task planners.
%They evaluate the system with experiments on teaching the robot to stack blocks, pour from a bottle and open a door.
However, the approach requires 5-10 demonstrations to learn action conditions which becomes tedious and impractical if several actions need to be taught.
Our work argues for having the user act as the expert by letting them correct inferred action conditions, thus allowing a new action to be learned from a single demonstration.
%and uses k-means and entropy to deduce action conditions from demonstrations. 
We further provide a graphical interface that allows users to create new actions and previously unseen problems that can be solved with task planners.

\section{APPROACH}
\label{sec:approach}
Our approach aims at providing end-users with an intuitive way of teaching robots new actions that can be reused with a task planner to solve more complex tasks.
Given a single demonstration, the robot should learn both \textit{how} (\sect{sec:lowlevel}) and \textit{when} (\sect{sec:highlevel}) an action should be applied.
To accelerate the programming process, action conditions are directly inferred from a single demonstration (\sect{sec:inference}).
The action generalisation is performed on both low- and high-level representations (\sect{sec:generalisation}), allowing it to be reused with a task planner (\sect{sec:planning}).
We will describe our approach in the following sections.

\subsection{Low-level Action Representation}
\label{sec:lowlevel}
We represent low-level actions as proposed in previous work using keyframe-based PbD \cite{alexandrova2014robot}, where the action is represented as a sparse sequence of gripper states (open/close) and end-effector poses relative to perceived objects or to the robot's coordinate frame.
During the demonstration phase the user guides the robot arm using kinesthetic manipulation and saves poses that they find relevant for the action.
For example, the pick-and-place action of an object to a marked position could be represented as poses relative to the object (for the pick action), poses relative to the target position (for the place action), and corresponding open/close gripper states. 
Action executions are performed by first detecting the landmarks in the environment, calculating the end-effector poses relative to the observed landmarks, and interpolating between the poses.

While these actions can be learned from multiple demonstrations \cite{niekum2012learning}, we take the approach that only requires a single demonstration by heuristically assigning poses and letting the user correct them if needed \cite{alexandrova2014robot}.
Thus, the first demonstrated action is already an executable action.
The user can teach multiple manipulation actions and discriminate between them by associating different conditions that specify \textit{when} the robot should use them (\eg actions using claw or suction grippers).
These conditions are discussed next in \sect{sec:highlevel}.
% In iRoPro, new actions are initialised with the robot's end-effector poses in a neutral position (as seen in \fig{fig:dispositif}) to allow unobstructed object detection.

% \begin{figure}
% \includegraphics[width=0.8\linewidth]{figures/object-type-hierarchy.png}
% \caption{Type hierarchy describing a general type `Element' that includes `Object' and `Position' and three object types `Base', `Cube', and `Roof'.}
% \label{fig:type-hierarchy}
% \end{figure}

  \begin{figure}[t]
   \centering
\includegraphics[width=\linewidth]{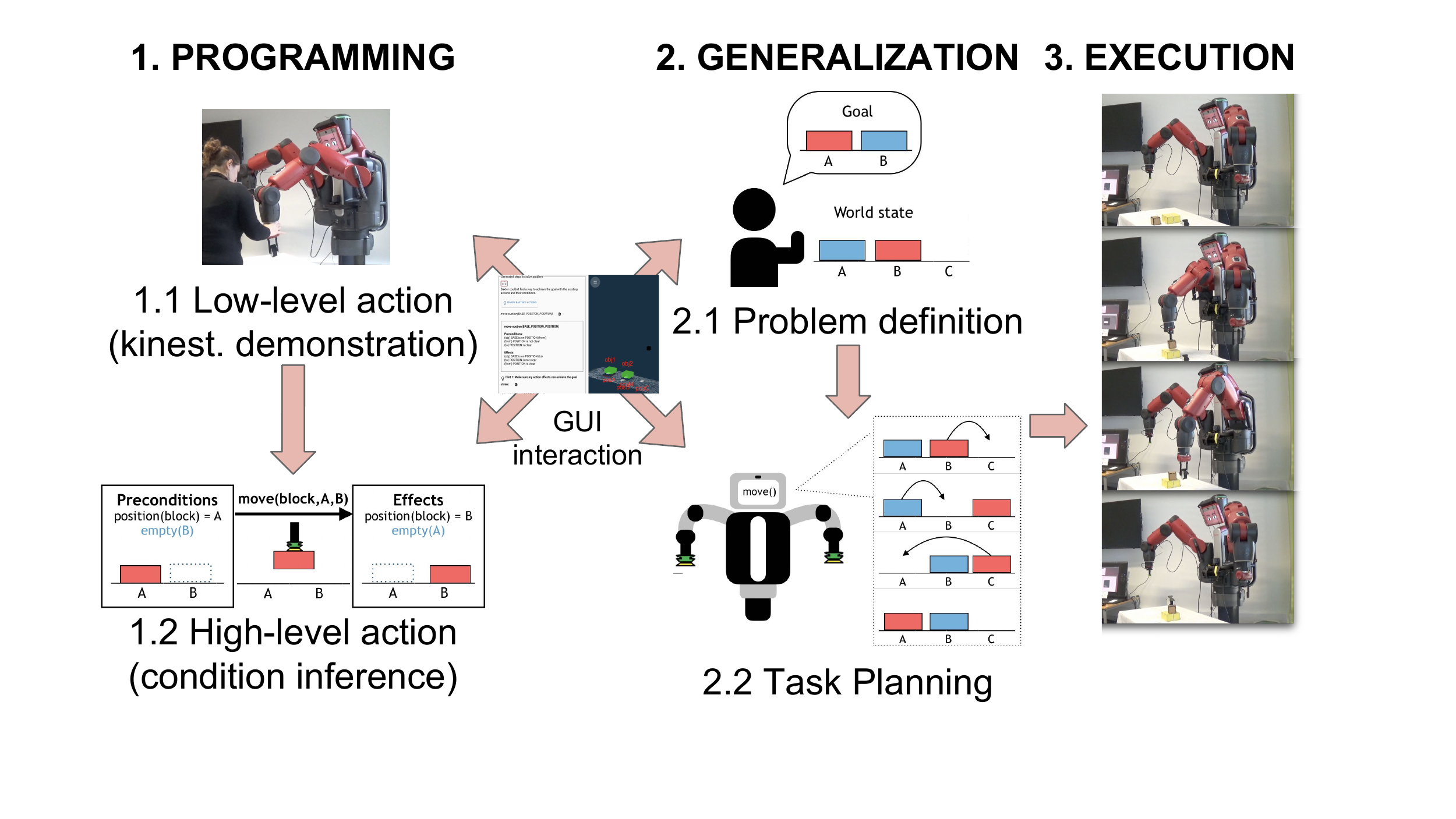}
   \caption{Overview of iRoPro to teach low- and high-level actions: the user interacts with the GUI to run the demonstration, modify inferred action conditions and to create new planning problems for the robot to solve and execute.}
   \label{fig:overview}
  \end{figure}

\subsection{High-level Action Representation}
\label{sec:highlevel}
We represent high-level actions similar to previous work on task planning \cite{ghallab2004automated}, where an action is represented as a tuple $a = (\text{param}(a), \text{pre}(a),$ $\text{eff}(a))$, whose elements are:
\begin{itemize}
\item $\text{param}(a)$: set of parameters that $a$ applies to
\item $\text{pre}(a)$: set of predicates that must be true to apply $a$
\item $\text{eff}(a)^{-}$: set of predicates that are false after applying $a$
\item $\text{eff}(a)^{+}$: set of predicates that are true after applying $a$
\end{itemize}
where $\text{eff}(a) = \text{eff}(a)^{-} \cup \text{eff}(a)^{+}$. 
Action parameters are world instances that the robot interacts with and are associated with a \textit{type}.
For example, in iRoPro, we implemented a type hierarchy, consisting of a general type ELEMENT, divided into POSITION and OBJECT, which further divides into BASE, CUBE, and ROOF (\fig{fig:dispositif}).

Predicates are used to describe object states and relations between them and are defined in first-order logic.
In our graphical interface, predicates are translated from first-order logic (`on(obj, A)') to English statements (`obj is on A').

In iRoPro, we implemented predicates that are commonly used in task planning domains as well as two additional ones to describe object properties:
\begin{itemize}
    \item \textit{ELEMENT is clear}: an element has nothing on top of it
    \item \textit{OBJECT is on ELEMENT}: an object is on an element
    \item \textit{OBJECT is stackable on ELEMENT}: an object can be placed on an element
    \item \textit{OBJECT is flat}: an object has a flat top
    \item \textit{OBJECT is thin}: an object is thin
\end{itemize}
We assume that CUBE and BASE objects are flat, while CUBE and ROOF objects are thin enough for the robot to grasp.
The set of inferred types and predicates could be extended for more complex tasks (\eg object colour or orientation \cite{li2016learning}).
% In our implementation we assume that CUBE and BASE objects are flat and that CUBE and ROOF objects are thin enough for the robot to grasp.
% \begin{tabular}{rl} \hline
% & \textbf{move(obj1, A, B):}\\ \hline
% Param: & obj1 - object, A - position, B - position\\
% Pre: & obj1 is on A, B is clear \\
% Eff$^{+}$: & obj1 is on B, A is clear\\
% Eff$^{-}$: & obj1 is not on B, B is not clear\\
%  & \\
% \end{tabular}

%An example of a move action of an object from position A to B is shown in \fig{fig:action-model}.
%Action conditions can be learned from multiple demonstrations (\citet{abdo2013learning}).
%We take an approach where the action can be learned from a single demonstration by inferring parameters, preconditions, and effects from observed states.
%We will discuss this in the next section.

\subsection{Action Inference from Demonstration}
\label{sec:inference}
Instead of manually defining action parameters, preconditions, and effects, we accelerate the programming process by inferring them from the observed sensor data during the teaching phase.
Object types are inferred based on their detected bounding boxes (see \sect{sec:platform}).
Object positions are determined by the proximity of the object to given positions.
If the nearest position \emph{p} to the object \emph{o} is within a certain threshold $d$, then the predicates `\emph{o} is on \emph{p}' and `\emph{p} is not clear' are added to the detected world state.

\begin{figure}
\centering
\includegraphics[width=0.9\linewidth]{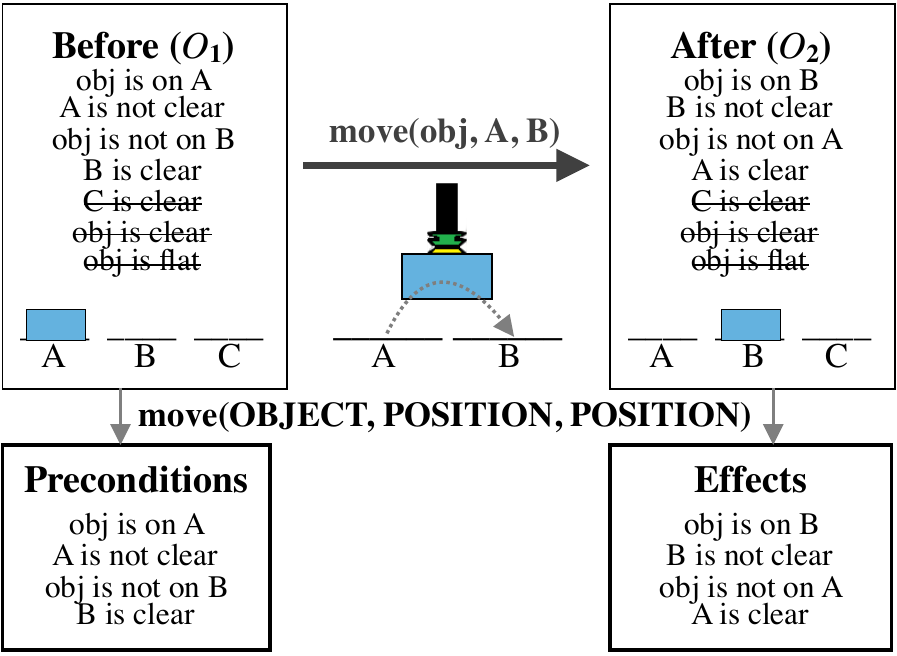}
\caption{Example of a high-level action for moving an object from A to B. Conditions are inferred from the observed predicates before ($O_1$) and after ($O_2$) the demonstration.
}
\label{fig:action-model}
\end{figure}

To infer action conditions, the robot perceives the initial world state   before %(\textit{precondition})
and after the kinesthetic action demonstration as seen in similar work for learning object manipulation tasks \cite{ahmadzadeh2015learning}.
%(\textit{effects}).
Let $O_1 = \{\phi_1, \phi_2, ... \}$ be the set of predicates observed before the action demonstration and $O_2 = \{\psi_1, \psi_2, ... \}$ after.
The action inference is the heuristic deduction of predicates that have changed between $O_1$ and $O_2$, \ie
\begin{align*} \text{pre}(a) = (O_1 - O_1 \cap O_2) = \{\phi_i | \phi_i \in O_1 \wedge \phi_i \notin O_2 \}, \\
\text{eff}(a) = (O_2 - O_1 \cap O_2) = \{\psi_i | \psi_i \notin O_1 \wedge \psi_i \in O_2 \}, 
\end{align*}
where $\text{eff}(a)$ includes positive and negative effects (\fig{fig:action-model}).
A predicate $\phi$ has variables $\text{var}(\phi) = \{v_1, v_2, \dots\}$, where each $v_i$ has a type.
Therefore, action parameters are the set of variables that appear in either preconditions or effects, \ie
\begin{align*}
     \text{param}(a) = \{v_i &| \hspace{0.3cm}\exists \phi \in \text{pre}(a) \text{ s.t. } v_i \in \text{var}(\phi)\\
     &\lor \exists \psi \in \text{eff}(a) \text{ s.t. } v_i \in \text{var}(\psi) \}.
\end{align*}
%where $type(p_i) = type(v_i)$.
%First infer the positions, then state that all other elements must be clear

Note that conditions could be learned from multiple demonstrations \cite{abdo2013learning,konidaris2018fromSkills}.
Our work argues for accelerating the teaching phase by learning from a single demonstration and letting the user act as the expert to correct wrongly inferred conditions.

\subsection{Action Generalisation}
\label{sec:generalisation}
% Taught actions can be customised to specific use cases and are generalised on both representation levels:
% Due to simultaneously learning low- and high-level action representations, an action is generalised in two ways:
The low-level representation (\sect{sec:lowlevel}) generalises motion trajectories by re-calculating poses based on detected landmarks from the demonstrated to the new environment.
The high-level representation (\sect{sec:highlevel}) specifies when an action can be applied, therefore allows taught low-level motion trajectories to be reused for other objects (\eg use suction grip for all objects, regardless of their dimension) or to be restricted for certain types (\eg only BASE objects).
By combining these two representation levels, taught actions can be generalised for more complex environments and the user can customise them for their specific use case.

\subsection{Task Planning}
\label{sec:planning}
Task planners are used to generate solutions, or action sequences, to solve complex problems.
%ught actions are reused with existing task planners by translating them into PDDL (\citet{mcdermott1998pddl}). 
Given a description of a planning \textit{domain}, \ie object types, actions with preconditions and effects, we can define a planning \textit{problem} with an initial state and a desired goal state. 
The planner generates an optimal action sequence, or \textit{plan}, which guarantees the transition from initial state to the goal state. 
PDDL \cite{mcdermott1998pddl} %extends the STRIPS \cite{fikes1971strips} formalism and 
is often used as a standard encoding language for planning problems.
%as it allows type structures
A move action as shown in \fig{fig:action-model} is defined as follows:

% \small{
\begin{verbatim}
(:action move
 :parameters (?obj - object 
              ?A - position ?B - position)
 :precondition (and (on ?obj ?A)(clear ?B)
            not(on ?obj ?B) not(clear ?A))
 :effect (and (on ?obj ?B) (clear ?A)
           not(on ?obj ?A) not(clear ?B))
 \end{verbatim}
% }
\vspace{-0.5cm}
A planning problem consists of an initial state and a goal state and can be solved using existing actions in the domain.
For example, the problem of swapping two objects obj1, obj2 on A and B respectively with C unoccupied, can be defined as:%\footnote{An example of a complete planning domain and problem in PDDL can be found in the attached materials}:
% \small{
\begin{verbatim}
(:objects obj1 obj2 - object
          A B C - position)
(:init (and (on obj1 A) (on obj2 B)
            (clear C))
(:goal (and (on obj1 B) (on obj2 A)))
 \end{verbatim}
 \vspace{-0.5cm}

%  \resizebox{0.5\textwidth}{!}{\theverbbox}

% \begin{verbatim}
% (:objects obj1 obj2 - object
%           A B C - position)
% (:init (and (on obj1 A) (on obj2 B) (clear C)))
% (:goal (and (on obj1 B) (on obj2 A)))
% \end{verbatim}
% }
The planner would generate the following action sequence:
% \small{
\begin{verbatim}
1. move(obj1, A, C)
2. move(obj2, B, A)
3. move(obj1, C, B)
\end{verbatim}
% }

\section{System}
\label{sec:system}
\subsection{Platform \& Implementation Details}
\label{sec:platform}
We implemented our system on a Baxter robot with two arms (one claw and one suction gripper), both with 7-DoF and a load capacity of 2.2kg.
For the object perception we mounted a Kinect Xbox 360 depth camera on the robot.
We developed a user interface as a web application that can be accessed via a browser on a PC, tablet or smartphone.
The source code for iRoPro is developed in ROS \cite{quigley2009ros} and available online\footnote{\url{https://github.com/ysl208/iRoPro/tree/cond}}. %ysl208/rapid\_pbd}.
The low-level action is learned using the open-source system Rapid PbD\footnote{\url{https://github.com/jstnhuang/rapid\_pbd}}.
%for teaching actions by keyframe-based demonstration.
The integration of the task planner is implemented using the ROS package PDDL planner\footnote{\url{http://docs.ros.org/indigo/api/pddl\_planner}}.

% \subsection{Implementation Details}
% \label{sec:implementation}
In our implementation, landmarks are either predefined table positions or 
objects that are detected from Kinect point cloud clusters using an open-source tabletop segmentation library\footnote{\url{https://github.com/jstnhuang/surface\_perception}}.
An object $obj = (x,y,z, width, length, height)$ is represented by its detected location and bounding box, which are used to infer its type and related predicates (\sect{sec:inference})
% We implemented a partial PDDL domain in iRoPro that includes a set of predefined object types and predicates (\sect{sec:highlevel}).
The user creates a complete PDDL domain via the GUI by teaching new actions and problems that can be solved with the integrated task planner.\footnote{Video can be viewed at \url{https://youtu.be/YCDrC0UFX38}}

% To that end, the user mainly interacts with the system for two main aspects: Actions and Problems.
% We will discuss them in the following sections.

\subsection{Interactive Robot Programming}
\label{sec:interactive}
\begin{figure}[t]
\includegraphics[width=\linewidth]{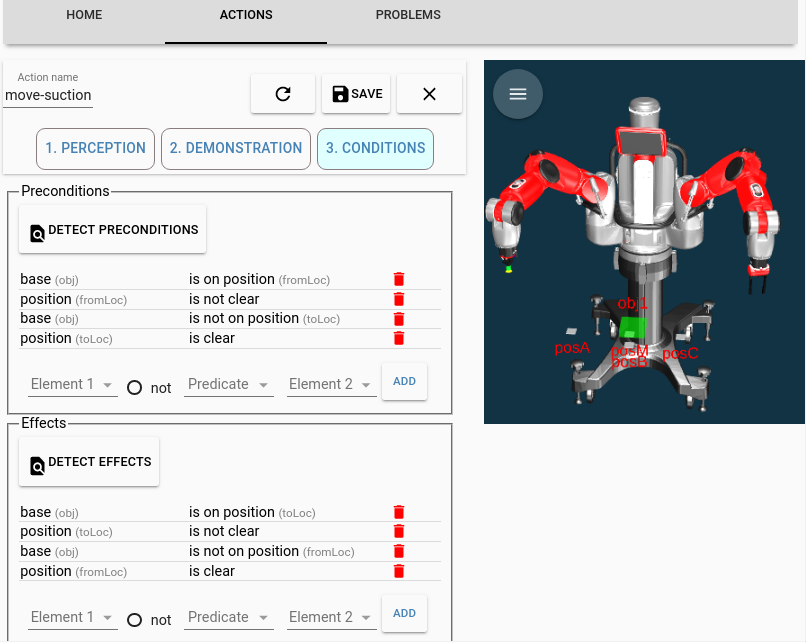}
    \caption{The iRoPro interface showing the action condition menu and an interactive visualisation of the Baxter robot and detected objects.}\label{fig:gui-action-3}%
\end{figure}

The user interacts with the GUI (\fig{fig:gui-action-3}) to visualise the robot and the detected objects, create new actions, run the kinesthetic teaching by demonstration, modify inferred types or predicates, create and solve new problems with the task planner.
The interactive robot programming cycle consists of creating and modifying actions and problems.
\paragraph{Actions} New actions are taught by kinesthetically moving the robot's arms, where both low-level and high-level actions are learned and generalised.
The low-level action is learned %using the open-source system Rapid PbD for teaching actions 
by keyframe-based demonstration (\sect{sec:lowlevel}).
To verify the taught action, the user can have the robot re-execute it immediately.
The high-level action is inferred automatically by capturing the world state before and after the action demonstration (as described in \sect{sec:inference}).
The user can modify the action properties if the inference was not correct.
% Action generalisation can be done by modifying action properties (\sect{sec:generalisation}).
To teach more actions, the user can either create a new one or copy a previously taught action and modify it.

\paragraph{Problems} New planning problems can be created if at least one action exists.
To create a problem, the robot first detects the existing landmarks and infers their types and initial states.
The user can modify them if the inference was not correct.
Then, the user enters predicates that describe the goal states to achieve.
The complete planning domain and problem are translated into PDDL and sent to the Fast-Forward planner \cite{hoffmann2001ff}.
If a solution is found that reaches the goal, it is displayed on the GUI for the user to verify and execute on the robot.
If no solution is found or if the generated plan is wrong, the user can open a debug menu which summarises the entire planning domain with hints described in natural language to investigate the problem (\eg `make sure the action effects can achieve the goal states').
In our user study (\sect{sec:quanteval}) we found that this helped users understand how the system worked and why the generated plan was wrong.
Once the user modified actions, initial or goal states, they can relaunch the planner to see if a correct plan is generated.
To solve new tasks, the user can create a new problem or modify existing ones by redetecting the objects.

\begin{figure*}[t]
\centering
 \includegraphics[width=0.9\linewidth]{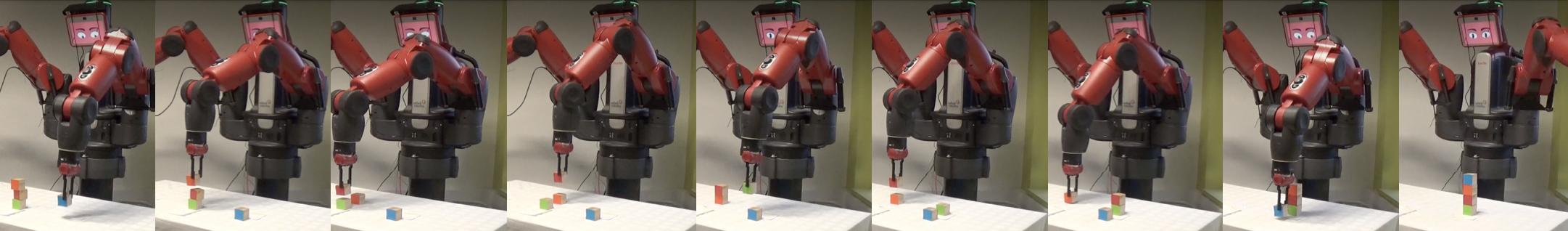}
 \includegraphics[width=0.9\linewidth]{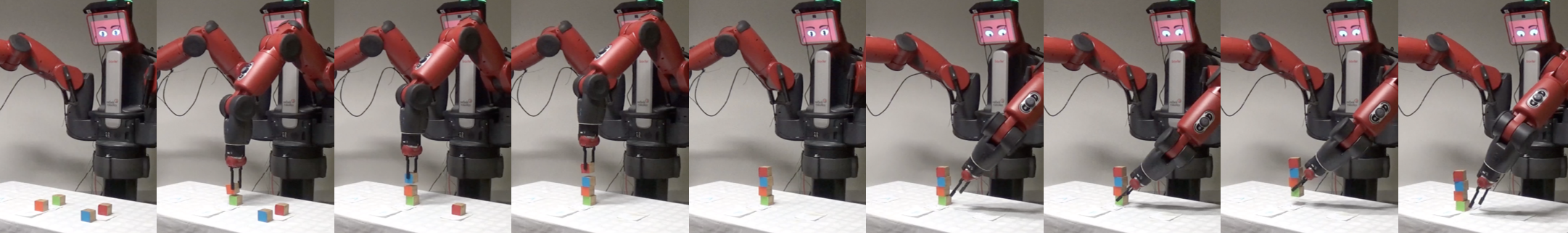}
\caption{Snapshots from the executions of the system evaluation showing Task 3 (top) and 4 (bottom).}
\label{fig:filmstrip}
\end{figure*}
\subsection{Plan Execution} 
The generated plan is a sequence of actions with parameters that correspond to detected objects.
For each action, the sequence of end-effector poses are calculated relative to the landmarks that the action is being applied to (\sect{sec:lowlevel}).
To accelerate the execution, we only detect the landmarks once at the start and save their new positions in a mental model.
After each action execution, the user can confirm that it executed correctly and the mental model is updated with the latest positions of the changed landmarks.
The mental model is also used as a workaround for our limited perception system for problems with stacked objects in their initial states (\sect{sec:discussions}).

%\paragraph{Reusable Actions} 
%Our system allows the robot to reuse previously taught actions for unseen problems with different objects and types.
%The user is only required to teach the robot simple atomic actions and the robot reuses them for complex tasks that require multiple subactions.
%Without having to explicitly teach the robot the action sequence, the robot can solve any user-defined goal autonomously using the planner.
%To address new tasks and problems, the user can simply change problem parameters or action conditions if the programmed manipulation action can be transferred.

% steps for using the system are:
% \begin{enumerate}
%     \item Create a new planning domain
%     \item {Create new actions
%     \begin{enumerate}
%         \item Detect landmarks (parameters and types)
%         \item Kinesthetic demonstration of action
%         \item Assign conditions (preconditions/effects)
%     \end{enumerate}}
%     \item {Create new planning problem
%     \begin{enumerate}
%         \item Detect landmarks (initial state)
%         \item Assign goal state
%         \item Verify generated plan and execute
%     \end{enumerate}}
% \end{enumerate}

% \subsection{Teaching Strategies}
% \label{sec:strategies}
% \todo {either create one action and try to generalise it as much as possible by changing parameter types and adding preconditions, or copy an action to have the same manipulation action but change the parameters for other specific tasks. 
% As creating an action from scratch takes the longest, it is most optimal to reuse demonstrated manipulation actions.}
%\newpage

\section{SYSTEM EVALUATION}
\label{sec:syseval}
We evaluate our system's generalisability on six benchmark tasks (Table \ref{table:task-list}) and show how taught primitive actions can be reused for complex tasks.
The tasks involved manipulating different object types on four marked positions with both claw and suction grippers.
We take the Blocksworld domain \cite{slaney2001blocks} for building and rebuilding stacked objects (Tasks 1-4) and an elaborate version of the Tower of Hanoi problem with different object types to build a `house' (Tasks 5\&6).
Instead of disks, we decided to use different object types (ROOF, CUBE, BASE), where BASE corresponds to the largest disk, followed by CUBE and ROOF.
The rules for stacking different objects still apply (\eg BASE cannot be stacked on top of ROOF or CUBE).
However, to demonstrate the generalisation of actions to diverse tasks, additional constraints are added as objects cannot be all manipulated in the same way.
The order of the tasks was given with increasing complexity, requiring the user to modify existing actions or teach new actions from scratch.
% to show reusability of more and more complex tasks we use the Tower of Hanoi problem with different number of disks, to show its applicability in real-world, we take an assembly task.
% The most common tasks for industrial robots are assembly, packaging, and material handling (such as polishing etc.) \footnote{https://blog.robotiq.com/the-3-most-common-tasks-delegated-to-robots-in-manufacturing}.
%Picking, Packing and Palletizing – Most products are handled multiple times prior to final shipping. Robotic picking and packaging increases speed and accuracy along with lowering production costs \footnote{https://www.jabil.com/insights/blog-main/ten-popular-industrial-robot-applications.html}
% Assembly and packaging tasks require prior planning in order to complete a task efficiently.
%Thus, we take an assembly task to build a ..and a packaging task with.

\subsection{Protocol}
The tasks were programmed by one of the authors, with the most efficient teaching strategy of minimising the number of actions created and generalising them by changing the action properties (as described in \sect{sec:generalisation}).
Depending on the given task and involved objects, the experimenter decided what manipulation action needed to be taught.
Only one planning problem was created and reused for all tasks by redetecting the objects in the initial state and changing the goal state.
When the generated plan was incorrect, the debug menu on the GUI was used to determine the changes to be made to generalise the actions.
A task was considered completed when the generated plan was correct and the robot successfully executed it.
As the mental model saved the latest object positions after an action execution, Tasks 3 and 6 were continued from the preceding tasks and did not require redetecting the initial states. 
%reusing the problem with the latest saved mental model allowed them to overcome the limitation to detect stacked objects.

\subsection{Results}
We programmed three manipulation actions for the six benchmark tasks, which involved demonstrating pick-and-place actions with claw and suction grippers from the \textit{top} and from the \textit{side}.
Actions were generalised by changing parameter types (\eg from CUBE or POSITION to ELEMENT) or adding preconditions or effects which were not inferred automatically.
For pick-and-place actions from the top, `obj is clear' was added as a precondition (Task 1-3), while it was not included when picking an object from the side to allow moving a pile of objects (Task 4).
For actions involving the claw gripper, the precondition `obj is thin' was added so that the robot would only use it on ROOF and CUBE objects, similarly `is flat' for the suction gripper (Task 5\&6).
The `is stackable' condition was used for the Tower of Hanoi as an equivalent to the rule `larger objects cannot be placed on top of smaller ones'.
Due to the noisy sensing and control of the Baxter robot, action executions failed occasionally, even though the generated plan was correct.
Overall, the robot was able to generate plans for all tasks and executed them at least twice (\fig{fig:filmstrip}).
While the experimental scope was limited and set in a controlled environment, it still demonstrates iRoPro's expressivity. 
With minimal end-user programming effort, manipulation actions can be taught from scratch and reused for a diverse range of tasks, even beyond the six benchmark tasks.
% This evaluation shows the generalisability of our system, allowing us to teach primitive actions by demonstration and reuse them with a task planner to solve more complex problems.

\begin{table}[t]
\begin{center}
\caption{Benchmark tasks for the system evaluation. Three different pick-and-place actions were programmed.}
\label{table:task-list}
%\begin{center}
\begin{tabular}{l@{\hskip1.9pt}l@{\hskip1.9pt}l}
\textbf{\#} & \textbf{Task goal} & \textbf{Pick-and-place action}\\ \hline
1 & Build tower with 3 CUBES & claw from top \\
2 & Build tower with 4 CUBES & claw from top \\
3 & Rebuild Task 2 on a different position & claw from top \\
4 & Build tower and move (w/o disassembly) & claw from top \& side \\
5 & Build house with BASE, CUBE, ROOF & claw \& suction from top \\
6 & Rebuild Task 5 on a different position & claw \& suction from top \\
% 5 & Assembly & P\&p with turning/orienting \\
% 6 & Packaging & P\&p with pushing to save space \\
\end{tabular}
\end{center}
\end{table}
\section{USER EVALUATION}
\label{sec:quanteval}
The second part of the evaluation was conducted using THEDRE \cite{mandran2017thedre}, a human experiment design method that combines qualitative and quantitative approaches to continuously improve and evaluate the developed system from the experimental ground.
%which aims to evaluate computer systems in a research context by integrating a user-centered approach.
% is based on continuous improvement and takes a pragmatic constructivist approach \cite{avenier2015finding}, allowing it to further develop the system as well as the scientific knowledge from the experimental ground.
% To that end, it offered us the possibility to mix qualitative and quantitative approaches in order to gather as much data as possible to evaluate and improve our system. 
The aim was to evaluate our approach with real end-users and we were also interested in the user's programming strategy for using the system.
We split participants into two control groups, with and without condition inference (\sect{sec:inference}) and evaluated user performance in terms of programming times for completing a set of benchmark tasks.
%if they had the tendency of `blindly trusting the system'.
% \subsection{Hypotheses}
We set the following hypotheses for our experiments:
\begin{enumerate}
    \item[H1] Action creation: users can teach new low- and high-level actions by demonstration
    \item[H2] Problem solving: users can solve new problems by defining the goal states and executing the plan on Baxter
    \item[H3] Autonomous system navigation: users understand the system and can navigate and troubleshoot on their own
    \item[H4] Condition inference (CI) - Group 1 vs 2: users without CI will understand the system better
    \item[H5] Pre-study test (PT): users that score higher in the PT have shorter programming times
\end{enumerate}

\subsection{Participants}
% The experiments were conducted with a Baxter robot with two grippers (suction and claw) and mounted with a Kinect Xbox 360 camera. 
% A table with 4 marked positions was placed in front of the robot with different objects to be manipulated by the robot during the experiment ().
% Participants had access to a computer with a mouse and a keyboard to program the robot via the graphical interface.

% \subsection{Participants}
The study was conducted with 21 participants (10M, 11F) in the range of 18-39 years (M=24.67, SD=6.1).
We recruited participants with different educational background and programming levels: 
6 `CS' (either completed a degree in computer science or were currently pursuing one),
7 `non-CS' (have previously taken a programming course before), 
and 8 `no experience' (only had experience with office productivity software).
Furthermore, 3 participants (in `CS') have programmed a robot before, out of which 1 had intermediate experience with symbolic planning languages while the remaining participants had no experience in either.
One participant in the category `non-CS' failed to complete the majority of tasks and was excluded from the result evaluation.
The two control groups included equal number of participants in all three categories.

\subsection{Protocol}
Users were first given a brief introduction to task planning concepts, the Baxter robot and the experimental set up (\fig{fig:dispositif}).
They were then asked to complete a pre-study test to capture the participant's understanding of the presented concepts.
Users were given 8 tasks to complete, where the first two were practice tasks to introduce them to the system (Table \ref{table:user_study_tasks}). 
The tasks were designed to address different aspects to familiarise them with the system:
create new actions (Task 6), modify parameter types (Tasks 4\&7), modify action conditions (Tasks 3,5,8).
For each task they needed to create a new problem, define the goal states, and launch the planner to generate an action sequence.
When the generated plan was correct, they were executed on the robot.
Otherwise, the user had to modify the existing input until the plan was correctly generated.
Tasks 6-8 were similar to the previous tasks (1-5) but use both robot grippers.

\subsection{Metrics}
We captured the following data during the experiments:
\begin{enumerate}
    \item \textbf{Qualitative data:} video recording of the experiment, observations during the experimental protocol.
    \item \textbf{Quantitative data:} task duration and UI activity log, pre-study test, post-study survey.
\end{enumerate}

The pre-study test included 7 questions related to their understanding of the concepts presented at the start of the experiment, \eg syntax (`If move(CUBE) describes a move action, tick all statements that are true.'), logical reasoning 
(`Which two conditions can never be true at the same time?'), and other concepts (`Tick all predicates that are required as preconditions for the given action').
The questions were multiple choice and the
%each question was normalised to count at most 1 point if answered correctly.
%The 
highest achievable score was 7.
%Answers that were selected incorrectly were penalised with half a point
%to discriminate users who selected unnecessary options

In the post-study survey we used the System Usability Scale (SUS) \cite{brooke2013sus} where participants had to give a rating on a 5-Point Likert scale ranging from `Strongly agree' to `Strongly disagree'.
It enabled us to measure the perceived usability of the system with a small sample of users.
As a benchmark, we compare overall responses to our previous user study \cite{liang2017evaluation}, where users were simulated a similar robot programming experience using the Wizard-of-Oz technique but had no direct interaction with a working system.
Finally, participants were asked which aspects they found most useful, most difficult, and which they liked the best and the least.%\footnote{Experimental documents can be found in the attached materials.}
% \footnote{The experimental documents used can be found in the attached material.}

% We compared the following data:
% \begin{itemize}
%     \item Participant profile (background, experience with programming/robots)
%     \item Pre-study questions on basic task planning concepts (i.e. predicate logic)
%     \item Post-study questionnaire on usability of the system
%     \item Notes during the experiment on user’s attempts/difficulties/etc.
%     \item UI activity log: timestamps of any button clicks
%     \item Number of tasks completed
%     \item Duration per completed task
% \end{itemize}

\begin{table}[t]
\centering
\caption{Benchmark tasks for the user study where the first two tasks were used to introduce participants to the system.}
\label{table:user_study_tasks}
\begin{center}
\begin{tabular}{l@{\hskip2.0pt}l}
\textbf{\# Task description} & \textbf{Main solution} \\ \hline
(1) move BASE object (suction grip) & create new action (+demo) \\
(2) move BASE object to any position & create new problem \\
 3 swap two BASE objects & add condition (`is clear') \\
 4 stack CUBE on BASE & modify types (`OBJECT')\\
 5 do not stack CUBE on ROOF & add condition (`is stackable')\\
 6 move ROOF object (claw grip) & create new action (+demo) \\
 7 stack ROOF on a CUBE & modify types (`ELEMENT') \\
 8 build a house (BASE, CUBE, ROOF) & navigate autonomously \\ \hline
\end{tabular}
\end{center}
\end{table}

\begin{table}[t]
\centering
\caption{User performance comparing task completion times with pre-study test scores.}
\label{table:performance}
\begin{center}
\begin{tabular}{r|rr|rr}
        & \multicolumn{2}{c}{\textbf{Main tasks (in min)}} & \multicolumn{2}{c}{\textbf{PT score (out of 7)}} \\
        & AVG                & STD                & AVG                   & STD                   \\ \hline
no experience  & 43.6               & 5.37               & 5.8                   & 0.95                  \\
non-CS  & 36.6               & 7.46               & 6.2                   & 0.55                  \\
CS      & 43.8               & 14.13              & 5.3                   & 1.11                 \\ \hline
\textbf{Overall} & \textbf{41.2}              & \textbf{9.08}               & \textbf{5.8}                   & \textbf{0.91}                  \\ \hline
Group 1 & 41.0               & 7.89               & 6.1                   & 0.77                  \\
Group 2 & 41.4               & 10.56              & 5.5                   & 0.98                 
\end{tabular}
\end{center}
\end{table}

% \begin{figure}[t]%
%   \centering
%     \includegraphics[width=0.8\columnwidth]{figures/quan-pretest-results.pdf}%
%     \caption{Participants who did better in the pre-study test completed the main tasks faster, with `non-CS' users scoring the highest and being the fastest on average.}\label{fig:pretestvstask}%
% \end{figure}%

\begin{figure*}[t]
  \includegraphics[width=0.98\linewidth]{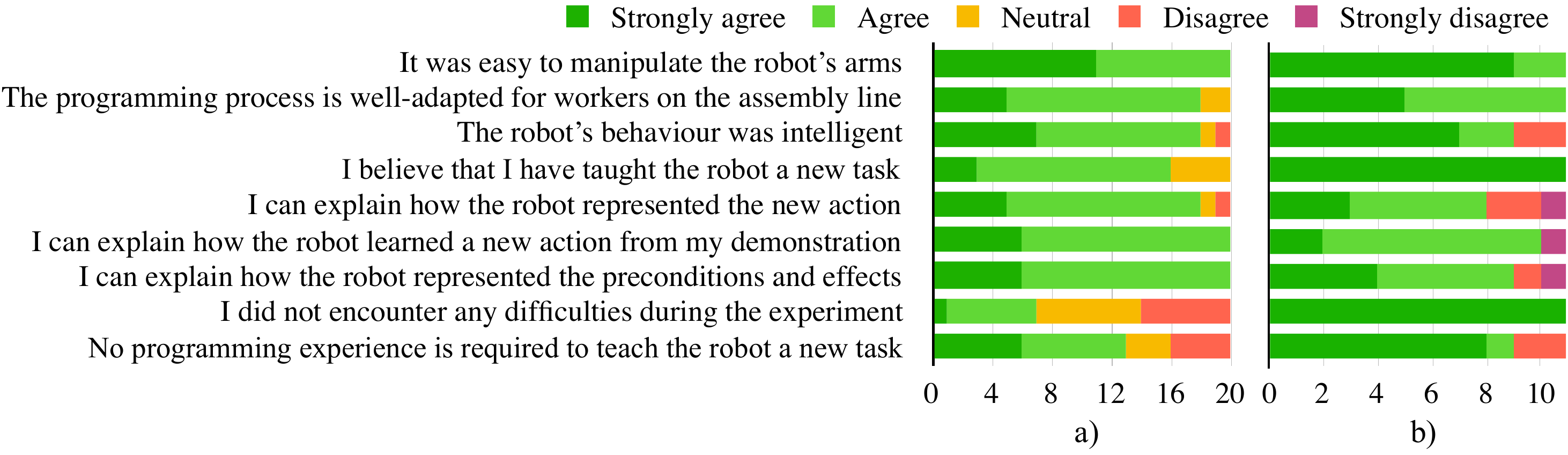}
  \caption{User responses from the post-study survey comparing a) iRoPro  (N=20) to b) our previous user study (N=11) \cite{liang2017evaluation}}
\label{fig:exp1vsexp2-results}
\end{figure*}
\subsection{Results}
20 participants completed all tasks, while one `non-CS' user failed to complete the majority of tasks and did not seem to understand the presented concepts.
This participant was excluded in the results presented below (Table \ref{table:performance}):

\subsubsection*{\textbf{H1)-H3) User performance}} 
%After the initial introduction to the system (Tasks 1\&2), all participants stated that they understood the difference between \textit{Actions} and \textit{Problems}.
Users took between 22-60 minutes to complete the main tasks (3-8), with an average of 41.2 minutes.
`non-CS' users completed the tasks the fastest, followed by users with no programming experience.
`CS' users took on average longer as they were often interested in testing the system's functionalities that were beyond the given tasks.

% \todo{check end time of last task is consistent}
%Task 1: Group 1: 7/10 participants did not modify the conditions and confirmed them directly. 
%
% \textbf{Task 2} (move BASE to any other position):
% 4 participants tried to set the goal as `obj is not on posM'. 
% This did not produce a plan as the action conditions were missing the right predicate.
Users initially had problems with different concepts that were presented at the start of the study, in particular they confused action parameters, preconditions and goal states.
For example, in Task 3, 6 (or 30\%) users tried to add intermediate action steps to achieve the goal state, instead of simply letting the planner generate the solution.
In Task 4, 14 (or 70\%) wanted to create a new action, even though they could reuse the existing action by modifying the parameter types.
However, by Task 6, all users were able to use the system autonomously to create new actions and problems and navigated the system with little to no guidance.
By the end of the experiment, users programmed two manipulation actions (one for each gripper) that were reused to complete all 8 benchmark tasks.
% The generated PDDL code for the planning domain can be found in the attached material.

% \begin{figure}[tb]%
%   \centering
%   \begin{subfigure}[t]{0.24\textwidth}%
%     \includegraphics[width=\textwidth]{figures/quan-pretest-results.png}%
%     \caption{Pre-test vs Task duration}\label{fig:pretestvstask}%
%   \end{subfigure}~~%
%   \begin{subfigure}[t]{0.24\textwidth}%
%     \includegraphics[width=\textwidth]{figures/quan-groups-duration.png}%
%     \caption{Task duration per group}\label{fig:taskvsgroup}%  
%   \end{subfigure}
%   \caption{}
%   \label{fig:task-duration}%
% \end{figure}%

% \begin{figure}[tp]
%   \subfigure[Pre-test vs Task duration]{\includegraphics[width=0.48\linewidth]{figures/quan-pretest-results.pdf}}
% \label{fig:pretestvstask}\quad
%   \subfigure[Task duration per group]{\includegraphics[width=0.48\linewidth]{figures/quan-groups-duration.png}}
% \label{fig:taskvsgroup}
% \end{figure}

\subsubsection*{\textbf{H4) Condition inference (CI)}} 
% To evaluate user programming strategies, we modified the action inference in our study to only infer a subset of predicates from the observed landmarks.
%, which did not cover those needed for later tasks.
We noticed a discrepancy in the programming strategies between the two control groups (Group 1 with CI vs. Group 2 without CI).
Participants in Group 1 had the tendency to leave the inferred conditions unmodified without adding conditions that would improve the action's generalisability to different use cases.
As participants in Group 2 had to add action conditions manually, they considered all predicates they deemed necessary for the action and added additional ones that were required for later tasks.
% On the other hand, 
%7/10 users in Group 1 did not modify the inferred conditions and did not put much thought potentially missing conditions.
Thus, Group 2 took on average longer to complete tasks where a new action had to be created (Tasks 1\&6), but was faster than Group 1 for subsequent tasks, where conditions had to be modified (Tasks 3,5,7).
Overall both groups had similar completion times for all tasks.
%(Group 1: AVG=41, STD=7.89 and Group 2: AVG=41.4, STD=10.56).

\subsubsection*{\textbf{H5) Pre-study test}} 
As expected, participants who demonstrated a better understanding of the introduced concepts in the pre-study test completed the main tasks (Tasks 3-8) faster on average (p-value$<0.05$). %(\fig{fig:pretestvstask}).
Users scored between 4.3-6.93 out of 7 points.
`non-CS' users scored above average points and completed the fastest.
As an outlier we observed that the fastest participant scored only 4.7, but easily learned how to use the system and completed the tasks in 22 minutes.
%the participant that took the longest (60min) scored 4.37.
Even though Group 1 performed slightly better in the pre-study test than Group 2, both completion times were on average similar.

%In Task 3: as all the users in Group 2 had added the missing condition (`posB is clear') in Task 1, a correct action sequence was generated directly.
%In Group 1 8/10 had not added this condition and needed guidance to resolve this issue, with 3 participants trying to modify the problem definition instead of the action, 1 trying to add a new action for `clearing a position', and 1 failing to figure out the right condition to add.

\subsubsection*{\textbf{System usability and learnability}} 
There are several ways to interpret the System Usability Scale (SUS) scores \cite{brooke2013sus} obtained from the post-study survey. 
Using Bangor et al.'s categories \cite{bangor2008suseval}, 14 (70\%) users ranked iRoPro as `acceptable', 6 (30\%) rated it `marginally acceptable', and no one ranked it `not acceptable'.
Correlating this with the Net Promoter Score, this corresponds to 10 (50\%) participants being `promoters' (most likely to recommend the system), 5 (25\%) `passive', and 5 (25\%) `detractors' (likely to discourage).
Overall, iRoPro was rated with a good system usability and learnability.

\subsubsection*{\textbf{Overall user experience}} 
%We asked users the same questions as in our previous work \textit{[Anonymous]} where we used the Wizard-of-Oz technique to conduct a qualitative evaluation of a potential system.
% similar with mainly positive responses in both studies .
We compare responses to our previous user study (N=11) \cite{liang2017evaluation}, where users had no direct interaction with the robot programming system as it was simulated using the Wizard-of-Oz technique.
The main differences were noted regarding difficulties encountered during the experiment (\fig{fig:exp1vsexp2-results}):
In our previous study we had 11 (or 100\%) agree that they encountered no difficulties, while this time only 7 (or 35\%) of our users stated the same.
However, all of our users claimed to have a good understanding of the action representation and how the robot learned new actions from their demonstrations, while an average of 2 (18\%) disagreed in \cite{liang2017evaluation}.
Both differences can be explained by the fact that in this study, users had to use an end-to-end system to program the robot, while in our previous work users had no direct interaction with a working system.
Even though our users encountered more difficulties, they got a better understanding of the functionalities due to getting hands-on experience.
This also correlates with negative responses in our survey to the question if `no programming experience was required' where 13/20 (65\%) agreed and 4 disagreed.
%3 out of the 4 who disagreed had at least taken a programming course before.
Overall, our user study received positive responses similar to the previous study.

 9 (45\%) users stated `generate solutions to defined goals automatically' as the most useful feature, followed by `robot learns action from my demonstration' (4 or 20\%) -- two main aspects of our approach.
4 (20\%) stated the most difficult part as `finding out why Baxter didn't solve a problem correctly', similarly 8 (40\%) stated difficulties related to `understanding predicates and defining conditions'. 
11 (55\%) disliked `assigning action conditions' the most, while the rest stated different aspects.
A common feedback was `it takes time to understand how the system works at the start'.
The most liked parts were `executing the generated plan' (8 or 40\%) and `demonstrating an action on Baxter' (7 or 35\%).
%Overall, users performed well in both control groups and gave positive feedback.

% \subsection{User strategies}
% As we have two control groups A (with condition inference) and B (without condition inference), we noticed two user strategies
% During the experiment, we noted the user's intended strategy to complete a task.

% \todo figure of people who tried the actions below:
% Teach new action
% Copy existing action
% Modify existing action

% \paragraph{User activity} 
% \todo {ignore some activity log at the end because of problems with executions} \\
% \todo {from Participant10 onwards, it keeps generating solutions for some reason}\\
% We logged the user activity on the interface at each button click to analyse the time spent on each task.

% The main activities 
% \todo {show graph of time spent at each option}

% \begin{figure}
% \includegraphics[width=0.9\linewidth]{figures/strategies}
% \caption{}
% \label{fig:strategies}
% \end{figure}

\section{DISCUSSIONS}
\label{sec:discussions}
In our evaluation scenarios we could have programmed more complicated manipulation actions such as turning or pushing for packaging tasks (as done previously in \cite{liang2018simultaneous}).
We decided to stick to simple pick-and-place manipulation actions, as our main focus was to evaluate iRoPro's usability with end-users. 
Both system and user evaluations demonstrated that the proposed robot programming process for manipulation tasks can be learned easily by users with or without programming experience.
% iRoPro minimises the user's programming overhead  can program a small set of primitive actions from scratch that can be reused to solve previously unseen problems with a task planner, we decided to stick to simple pick-and-place manipulation actions.

The workaround based on mental models of the environment only works if the environment is static and no external entity interferes with the world.
The next step would be to move from controlled environments to more dynamic ones, whereby all relevant aspects of the environment need to be fully perceivable by the system. 
An improved perception system would also allow tracking and verifying action executions in case of failures in more complex environments.

Due to the Baxter robot's different grippers, we did not program actions that use both arms simultaneously (\eg for carrying a tray). 
A possible extension would be to include a better motion and task planning system to allow this while also considering self-collision avoidance.
Furthermore, we did not program human-robot collaborative tasks, such as human-robot hand-over tasks.
To allow this, more complex planning domains and better multi-modal communication would need to be implemented.

%Users particularly liked the PbD technique and found that
%`generating plans automatically' was the most useful feature 
% \begin{enumerate} 
%Our object perception is limited as it does not detect objects that are too close together (\eg stacked objects).

% As mentioned in \sect{sec:implementation}, we partly the latter by using a mental model, where we first executed a stacking task to save the latest object positions, then reused the saved state for subsequent tasks.

% We only included a minimal set of predicates that we deemed intuitive and useful for object manipulation tasks. To capture more complex domains such as object orientation \cite{li2016learning}.
%the use a mobile robot that can move between workstations.
%\item A possible extension would be to incorporate probabilistic techniques to learn predicates or pre-train the robot on simulated scenarios to improve the condition inference.
% \end{enumerate}

\section{CONCLUSION} 
\label{sec:conclusion}
% \addtolength{\textheight}{-1cm}   % This command serves to balance the column lengths
In this work we presented iRoPro, an interactive Robot Programming system that allows simultaneous teaching of low- and high-level representations of actions by demonstration.
The robot reuses the actions with a task planner to generate solutions to previously unseen tasks that are more complex than the demonstrated action.
The approach was implemented on a Baxter robot and we showed its generalisability on six benchmark tasks by teaching a minimal set of primitive actions that were reused for all tasks.
We further demonstrated its usability with a user study (20 successful results, 1 unsuccessful result) where participants with diverse educational backgrounds and programming levels learned how to use the system in less than an hour.
Both user performance and feedback confirmed iRoPro's usability, with the majority ranking it as `acceptable' and half being promoters.
Overall, we demonstrated that our approach allows users with any programming level to efficiently teach robots new actions that can be reused for complex manipulation tasks.

Future work will focus on exploring more challenging domains to extend the work to other platforms by including a wider range of predicates and probabilistic techniques to improve the condition inference.
As we focused on controlled environments, further studies will involve more complex environments with factory workers who may ultimately use this technology.




% \section*{ACKNOWLEDGMENT}

% The preferred spelling of the word ÒacknowledgmentÓ in America is without an ÒeÓ after the ÒgÓ. Avoid the stilted expression, ÒOne of us (R. B. G.) thanks . . .Ó  Instead, try ÒR. B. G. thanksÓ. Put sponsor acknowledgments in the unnumbered footnote on the first page.

\section*{Acknowledgments}
We would like to thank Nadine Mandran for supporting the experiment design process.
% and Justin Huang for providing the open-source system Rapid PbD and his continuous support, and .
% Put sponsor acknowledgments in the unnumbered footnote on the first page.

%% Use plainnat to work nicely with natbib. 
\newpage
\bibliographystyle{IEEEtran}
\bibliography{references}

\end{document}